\documentclass{article}

\usepackage[final]{neurips_2022}

\usepackage[algo2e]{algorithm2e} 

\usepackage[utf8]{inputenc} 
\usepackage[T1]{fontenc}    
\usepackage{hyperref}       
\usepackage{url}            
\usepackage{booktabs}       
\usepackage{amsfonts}       
\usepackage{nicefrac}       
\usepackage{microtype}      
\usepackage{xcolor}         
\usepackage{booktabs}
\usepackage{xspace}
\usepackage{graphicx}
\usepackage{amsmath}
\usepackage{algorithm}
\usepackage{algorithmic}
\usepackage{caption}
\usepackage{subcaption}

\usepackage{amsmath,amsfonts,bm}

\def\eqref#1{equation~\ref{#1}}

\def\1{\bm{1}}

\DeclareMathAlphabet{\mathsfit}{\encodingdefault}{\sfdefault}{m}{sl}
\SetMathAlphabet{\mathsfit}{bold}{\encodingdefault}{\sfdefault}{bx}{n}

\def\sP{{\mathbb{P}}}

\usepackage{wrapfig}
\newcommand{\model}{E-MAPP\xspace}
\usepackage{tabularx}
\usepackage{subcaption}
\usepackage{caption}
\usepackage{array}
\usepackage{natbib}
\setcitestyle{numbers,square}

\definecolor{MyDarkBlue}{rgb}{0,0.08,1}
\definecolor{MyDarkGreen}{rgb}{0.02,0.6,0.02}
\definecolor{MyDarkRed}{rgb}{0.8,0.02,0.02}
\definecolor{MyDarkOrange}{rgb}{0.40,0.2,0.02}
\definecolor{MyPurple}{RGB}{111,0,255}
\definecolor{MyRed}{rgb}{1.0,0.0,0.0}
\definecolor{MyGold}{rgb}{0.75,0.6,0.12}
\definecolor{MyDarkgray}{rgb}{0.66, 0.66, 0.66}

\newcommand{\myitem}{\vspace{-2pt}\item}

\title{\model: Efficient Multi-Agent Reinforcement Learning with Parallel Program Guidance}

\author{
Can Chang$^{1,2}$, \quad
Ni Mu$^{3}$, \quad
Jiajun Wu$^{4}$, \quad
Ling Pan$^{5}$, \quad
Huazhe Xu$^{1,2}$\\
\vspace{0.1cm}
$^{1}$IIIS, Tsinghua University $^{2}$Shanghai Qi Zhi Institute  $^{3}$Southeast University \\ $^{4}$Stanford University $^{5}$Mila, Universit\'e de Montr\'eal\\
\texttt{cc22@mails.tsinghua.edu.cn,
xuhuazhe12@gmail.com}
\vspace{1.0cm}
}

\begin{document}

\maketitle

\begin{abstract}
A critical challenge in multi-agent reinforcement learning~(MARL) is for multiple agents to efficiently accomplish complex, long-horizon tasks. 
The agents often have difficulties in cooperating on common goals, dividing complex tasks, and planning through several stages to make progress. 
We propose to address these challenges by guiding agents with programs designed for parallelization,
since programs as a representation contain rich structural and semantic information, and are widely used as abstractions for long-horizon tasks. 
Specifically, we introduce \textbf{E}fficient \textbf{M}ulti-\textbf{A}gent Reinforcement Learning with \textbf{P}arallel \textbf{P}rogram Guidance~(\model), a novel framework that leverages parallel programs to guide multiple agents to efficiently accomplish goals that require planning over $10+$ stages. 
\model integrates the structural information from a parallel program, promotes the cooperative behaviors grounded in program semantics, and improves the time efficiency via a task allocator. 
We conduct extensive experiments on a series of challenging, long-horizon cooperative tasks in the \emph{Overcooked} environment. Results show that \model outperforms strong baselines in terms of the completion rate, time efficiency, and zero-shot generalization ability by a large margin.
\end{abstract}

\section{Introduction}
Multi-agent reinforcement learning~(MARL) has achieved significant progress by advancing the cooperation of multiple agents to accomplish complex tasks, e.g., multi-robot control~\citep{jestel2021obtaining}, autonomous driving~\citep{vinitsky2018benchmarks, pmlr-v155-zhou21a}, and video games~\citep{wu2021too,arulkumaran2019alphastar}. Most recent advances in MARL focus on tasks that feature behavior coordination~\citep{samvelyan2019starcraft} or joint motion planning~\citep{semnani2020multi}. However, for \textit{long-horizon tasks} such as preparing a dish, 
existing methods often suffer from the inability to understand the task compositionality and subtasks' dependencies, resulting in inefficient cooperation and frequent conflicts. 
Therefore, a natural question to ask here is how we can solve long-horizon tasks in MARL, in the face of large state/action spaces and sparse feedback.

Long-horizon tasks are usually blessed with rich structure, and thus can be divided into a sequence of subtasks that can be resolved separately. 
Previous work~\citep{sun2019program,zhao2021proto} has introduced programs as instructions to help a single agent understand the task hierarchy and accomplish the task. Inspired by them, we develop a general multi-agent framework, where agents can leverage programs for accomplishing long-horizon tasks together. This is a challenging problem, and three substantial issues will emerge if multiple agents are naively enforced to follow sequential programs: first, sequential programs do not explicitly express the dependencies among subtasks, thus hindering the division of jobs among agents; second, different agents might have different abilities to accomplish certain subtasks or lines of programs; third, when assigned to a subtask together, multiple agents might need to collaborate without blocking resources with each other. 

As modern CPUs dispatch instructions to parallel processors, we propose a new multi-agent framework, \textbf{E}fficient \textbf{M}ulti-\textbf{A}gent Reinforcement Learning with \textbf{P}arallel \textbf{P}rogram Guidance~(\model), guiding cooperation and execution by automatically inferring the structure of parallelism from programs. Specifically, we first design a domain-specific language~(DSL) for multi-agent cooperation, and use multi-stage learning to ground subroutines of a given program into the agents' policy. Then, we learn feasibility functions, which entail the ability of agents to complete specific subroutines in the program in the status quo. Finally, we leverage the learned task structure to automatically enforce cooperation and division of labor among agents.

We conduct experiments on gradually more difficult challenges in the \textit{Overcooked}~\citep{wu2021too} environment. The \textit{Overcooked} environment requires the agents to cooperate on very long-horizon tasks, such as preparing dishes, while avoiding conflicting behaviors. 
Our method significantly outperforms other strong baselines  in completion rates and efficiency. The program structure also enables \model 
to deliver superior compositional generalization to novel scenes.

Our main contributions can be summarized as follows:
\begin{itemize}
\item We formulate a novel task of learning multi-agent cooperation via the guidance of parallel programs.
\item We present a novel framework for program grounding and long-horizon planning and instantiate the framework into practical multi-agent reinforcement learning algorithms.
\item We demonstrate the effectiveness of E-MAPP in completion rates and generalization ability over existing strong baselines and provide empirical analysis in long-horizon tasks.
\end{itemize}

\newcommand\paragraphtitle[1]{ \par\textbf{#1.}\quad}

\section{Related Work}
\paragraphtitle{Cooperative Multi-Agent Reinforcement Learning}
In a multi-agent cooperative game, agents collaborate with each other on a common goal~\citep{DBLP:journals/corr/abs-1908-03963}. Researchers have investigated many ways to facilitate agent coordination~\cite{jiang2018learning, ding2020learning, zhou2020smarts, peng2017multiagent, jaques2019social}. Value-based MARL algorithms engage in discovering the relationship between global value function and local value functions~\citep{sunehag2018value,rashid2018qmix}, while policy-based MARL algorithms use a centralized critic~\citep{yu2021surprising,lowe2017multi} to coordinate agent behaviors. More specifically, MAPPO~\citep{yu2021surprising} and MADDPG~\citep{lowe2017multi} leverage a fully-observable central critic to solve the issue of non-stationarity~\citep{DBLP:journals/corr/abs-1908-03963}. Value factorization approaches~\citep{rashid2018qmix,sunehag2018value,rashid2020weighted} decompose the global value function into a combination of local agent-wise utilities to cope with scalability, while policy factorization approaches~\citep{jain2020cordial} factorize the joint action space to coordinate marginal policies.

\paragraphtitle{Reinforcement Learning for Long-horizon Tasks} 
Reinforcement learning agents usually lack the ability to plan and reason over a long time horizon due to sparse rewards~\cite{wang2020long, emmons2020sparse,  savinov2018semi, eysenbach2019search, gupta2020relay, nasiriany2022maple}.
Goal-conditioned reinforcement learning~\citep{plappert2018multi} is one of the popular paradigms to address the sparse supervision problem. Imitation learning~\citep{ho2016generative,gupta2020relay} is another approach to solving the sparse reward problem. Another line of work has presented automatic goal generation and selection algorithms~\cite{andrychowicz2017hindsight,pitis2020maximum,li2020solving,davchev2021wish}; however, this introduces new challenges to design a suitable goal space that enjoys rich semantic meanings~\citep{dwiel2019hierarchical}. By contrast, our work uses the subtasks corresponding to possible subroutines in a program as goals, which are associated with domain knowledge. Hierarchical reinforcement learning~\citep[HRL; ][]{Sutton1998,singh2010intrinsically,mannor2004dynamic,stolle2002learning,NEURIPS2018_e6384711} is another path to solve long-horizon tasks, using a high-level policy for long-term planning and low-level policies for motion planning or specific behaviors. While our work is related to multi-agent hierarchical reinforcement learning~\citep{makar2001hierarchical}, which explicitly provides the directed acyclic task graph and the necessity of cooperation of each subtask, we focus on learning task structure such as subtask dependencies, loops, and branchings from the program and judging the necessity of cooperation without additional information.

\paragraphtitle{Instruction-Guided Agents}
Many recent advances have testified the advantages of leveraging structured prior knowledge such as task graphs~\citep{pmlr-v164-agia22a,jothimurugan2021compositional}, natural languages~\citep{saycan2022arxiv,chen2011learning,kaplan2017beating}  and programs~\citep{sun2019program,zhao2021proto} to promote efficient policy learning.
In contrast to other structured priors, programs stand out because of their strictly formatted and composable subroutines~\citep{bunel2018leveraging,chen2018execution}. Previous works leverage programs to enable a single agent to learn complex tasks by following programs~\citep{sun2019program,zhao2021proto}. However, a plethora of new challenges have been introduced, including task dependencies, collaboration schemes, and others; hence, it is difficult to trivially extend the existing works in the MARL settings. 
To enable multi-thread orderless policy execution, we leverage “plug-and-play” auxiliary functions to infer the relationships among subtasks.

\section{Problem Statement}
\begin{figure}[t] 
\centering
\includegraphics[trim=0 130 0 110, width=1.0\textwidth]{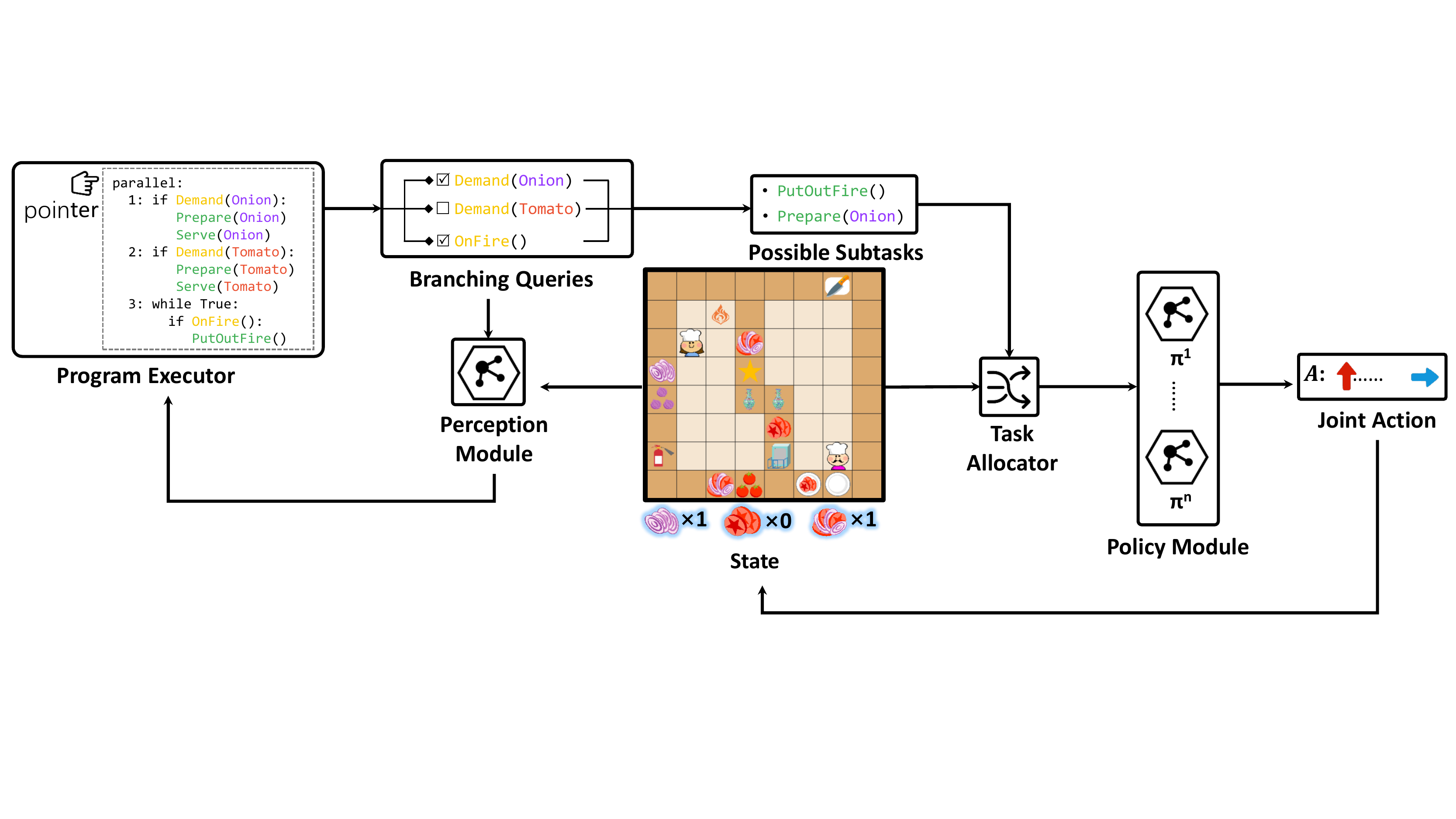}
\caption{\textbf{The overall framework of E-MAPP.} E-MAPP includes four components: 1)~A perception module that maps a query $q$ and the current state $s$ to boolean responses. 2)~A program executor that maintains a pool of possible subtasks and updates them according to the perceptive results. 3)~A task allocator that chooses proper subtasks from the subtask pool and assigns those to agents. 4)~A policy module that instructs agents in taking actions to accomplish specific subtasks. }
\label{framework}
\end{figure}
\subsection{Program Guided Cooperative Markov Game}

An infinite-horizon Markov Game is defined by a tuple $(\mathcal{N},\mathcal{S},\mathcal{A},\mathcal{T},\mathcal{R},\gamma)$, where $\mathcal{N}=\{1,\dots,N\}$ denotes the set of $N$ engaging agents, $\mathcal{S}$ denotes the state space, $\mathcal{A}=\mathcal{A}^1\times \dots \times \mathcal{A}^N $ denotes the Cartesian product of all the $N$ agents' action space, $\mathcal{T}$: $\mathcal{S}\times \mathcal{A}\rightarrow \mathcal{S}$ denotes the state transition function from current state $s$ to the next state $s'$ for the joint action $a=(a^1,\dots,a^N)$, $\mathcal{R}=\mathcal{R}^1 \times \dots \times \mathcal{R}^{N}$ denotes the Cartesian product of all the $N$ agents' reward functions, where each $\mathcal{R}^i$ determines the immediate reward for the $i$-th agent from the current state $s$ and joint action $a$, and $\gamma$ is the discount factor.

At each time $t$, each agent $i$ observes the current state $s_t$, makes the decision $a_t^i$, and receives the reward $r_t^i$. In a cooperative game, the collective goal is to optimize the joint policy $\mathbb{\pi}:\mathcal{S}\rightarrow A$ to maximize the sum of each agent's expected cumulative discounted rewards $\mathbb{E}_{\mathbf{a}_{t}^{i} \sim \pi^i\left(\cdot \mid \mathbf{s}_{t}\right),\mathbf{s}_{t} \sim \mathcal{P}}\left[\sum_{t=1}^{\infty}\sum_{i=1}^{N} \gamma^{t} r_t^i\left(\mathbf{s}_{t}, \mathbf{a}_{t}^{i}\right)\right]$.

A program-guided Markov game is a Markov game with a factorized state space. Specifically, state $\mathcal{S}=\mathcal{S}_e\times \mathcal{S}_p$, where $\mathcal{S}_e$ is the common state space of the environment and $\mathcal{S}_p$  is the multi-pointer program space~(see \hyperref[sec:multi-pointer program]{Section~\ref{sec:mpp}}). Accordingly, the state transition function takes as input the current compounded states $(s_e,s_p)$ and joint actions $a$ and then returns the next joint state $(s_e',s_p')$. In this study, the transition function of the program space is based on predefined rules.

\subsection{Parallel Programs} \label{sec:mpp}
\label{sec:multi-pointer program}
The program space consists of three components: a domain-specific language~(DSL) that contains all the possible subroutines, a set of pointers that point to relevant subroutines, and a control flow that manages the pointers.  A complete specification of the DSL used by our framework is in the Appendix~\ref{app:dsl}. Inside this DSL, a subroutine is a minimal executable unit in the program that corresponds to a subtask in the domain (e.g., \emph{Chop(Tomato)}). As in previous works~\citep{sun2019program,zhao2021proto}, we consider two types of subroutines: perception primitives (e.g., \emph{IsOnFire()}), which query about the status of the environment; and behavior primitives (e.g., \emph{Chop(Tomato)}, 
which issue an instruction. Control flow involves branching statements~(\emph{if/else}), loops~(\emph{for/while}), and \textbf{parallelism indicators}~(\emph{repeat/parallel}). The parallelism indicators are designed for multi-thread execution. The subroutines in a \emph{parallel} block are possible but not guaranteed to run concurrently, i.e., the agents should reason about what subroutines in a \emph{parallel} block can be executed concurrently. The subroutine in a \emph{repeat} block can be executed many times simultaneously.
We summarize all the current subroutines that are possible and executable as the \textit{Possible Subroutine Set}. The set of pointers points to all the subroutines in this set.

\subsection{Multi-Agent RL with Parallel Programs}
In our study, we aim to develop a framework for optimizing the joint policy in a parallel program-guided cooperative Markov game. To this end, the agents must keep track of the pointers in the program, learn to reason the primitives to choose the right branches, and  perform the action either collaboratively or individually to pursue high efficiency.

\section{Method}
\begin{figure}[t] 
\centering
\includegraphics[trim=20 260 220 0, width=0.9\textwidth, clip]{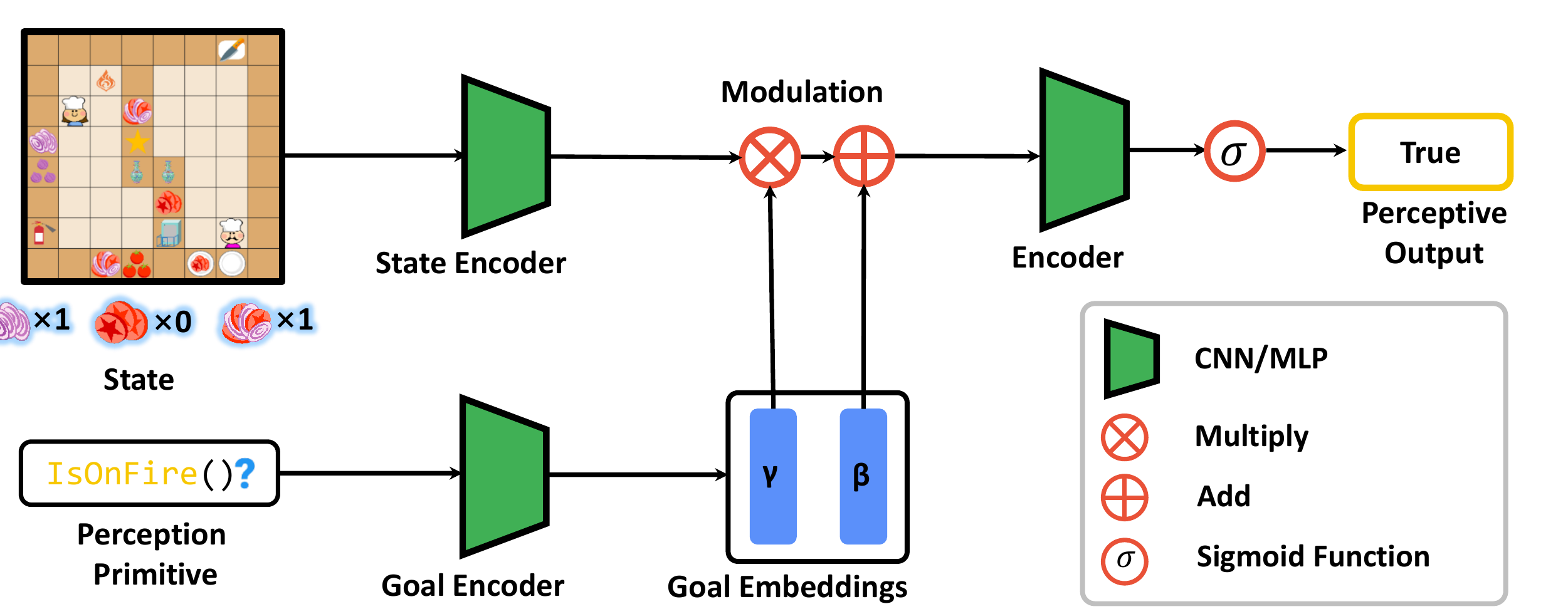}
\caption{\textbf{The perception module}. The perception module encodes the  perception primitive into modulation parameters, which operate on the state embeddings to get the goal-conditioned embeddings. These embeddings are then fed into another encoder, followed by a sigmoid function. The output is a real number in $[0,1]$, and will be binarized for branching selection by the program executor.}
\label{fig:perception module}
\end{figure}
Our goal is to enable multiple agents to cooperate to solve long-horizon tasks guided by parallel programs. There are three important factors we should consider: 
first, the agents should learn subtask-conditioned policies that can be composed together to accomplish long-horizon tasks and further generalize compositionally to unseen tasks; second, the agents should reason about task dependencies so that they can automatically parallelize tasks among them;
and third, the agents should distinguish between cooperative tasks and non-cooperative tasks that can be achieved by a single agent to avoid competition for common resources.

We propose \textbf{E}fficient \textbf{M}ulti-\textbf{A}gent Reinforcement Learning with \textbf{P}arallel \textbf{P}rograms~(\model), a multi-agent reinforcement learning framework where agents can cooperate to solve long-horizon tasks by following program guidance. As shown in Figure~\ref{framework}, \model includes four components: a perception module that is able to judge whether the queried status exists in the state; a parallel program executor that keeps track of subroutines and updates them based on the perception module output; a task allocator, which extracts queries of interest and obtains the corresponding set of possible subtasks from the program executor, and assigns tasks to agents; and a multi-agent policy module, which agents use to make their decisions based on the input subtask. 
The following of this section will introduce these four key components of  \model. The complete algorithm is shown in Appendix~\ref{app:algo}.

\subsection{Parallel Program Executor}
The program executor keeps a set of pointers pointing to possible subroutines in a domain-specific \emph{Possible Subroutine Set}. There are four types of control flows: \textit{if}-routine, \textit{while}-routine, \textit{parallel}-routine, and \textit{repeat}-routine. Meanwhile, there are two types of subroutines in our program: behavior primitives and perception primitives. 
\paragraphtitle{Control flows} An \textit{if}-routine contains a condition statement~(usually a perception primitive) and subroutines in the corresponding blocks. A  \textit{while}-routine contains a condition statement and a looping block of subroutines. A \textit{parallel}-routine contains parallel blocks of subroutines that are possible to be executed simultaneously. A \textit{repeat}-routine contains an unconditioned block of subroutines that can be executed multiple times by different agents. 
\paragraphtitle{Subroutines} A behavior primitive~(e.g., \emph{Chop(Tomato)}) corresponds to a subtask that must be completed by the agents. As for a perception primitive~(e.g., \emph{IsOnFire()}), it is a query that requires a boolean response. 

After an action is performed or a response to a perceptive query is received, the program executor updates its pointers and the Possible Subroutine Set. Detailed updating rules are shown in Appendix~\ref{app:mpu}. We note that if an achieved subtask does not correspond to any subroutine in the Possible Subroutine Set, the program executor will terminate the program immediately. This guarantees that no exceptions will occur due to violating the instructed order of subtasks.

\subsection{Perception Module}
\label{sec: perception module}
The perception module learns to map a query $q$ and the current state $s_t$ to a boolean answer $h=\phi(q,s_t)$. For example, when a perception primitive \emph{IsOnFire()} is passed to the perception module, it learns to check the existence of fire in the environment and returns true/false. The architecture is shown in Figure~\ref{fig:perception module}{}. Specifically, we randomly sample states, queries, and the ground-truth perception $h_{gt}$ as the training dataset and train the network $\phi$ in a supervised manner. We use binary cross entropy (BCE) loss $\mathcal{L}_{\text{perception}}=BCELoss(h_{\text{pred}},h_{\text{gt}})$ as the objective, where $h_{\text{pred}}$ denotes the perception output. In terms of architecture, we encode the perception primitive $s_p$ and the common state $s_e$ with a neural encoder. The encoded primitives are represented as $\gamma$ and $\beta$ to modulate the encoded common state. At the last layer, we use a sigmoid function to obtain the binary output. Training details and detailed architecture descriptions can be found in Appendix~\ref{app:architecture}. In this way, this module can determine whether the queried primitive exists in the state to aid the agents' decision-making.

\begin{figure}[t] 
\centering
\includegraphics[width=0.9\textwidth]{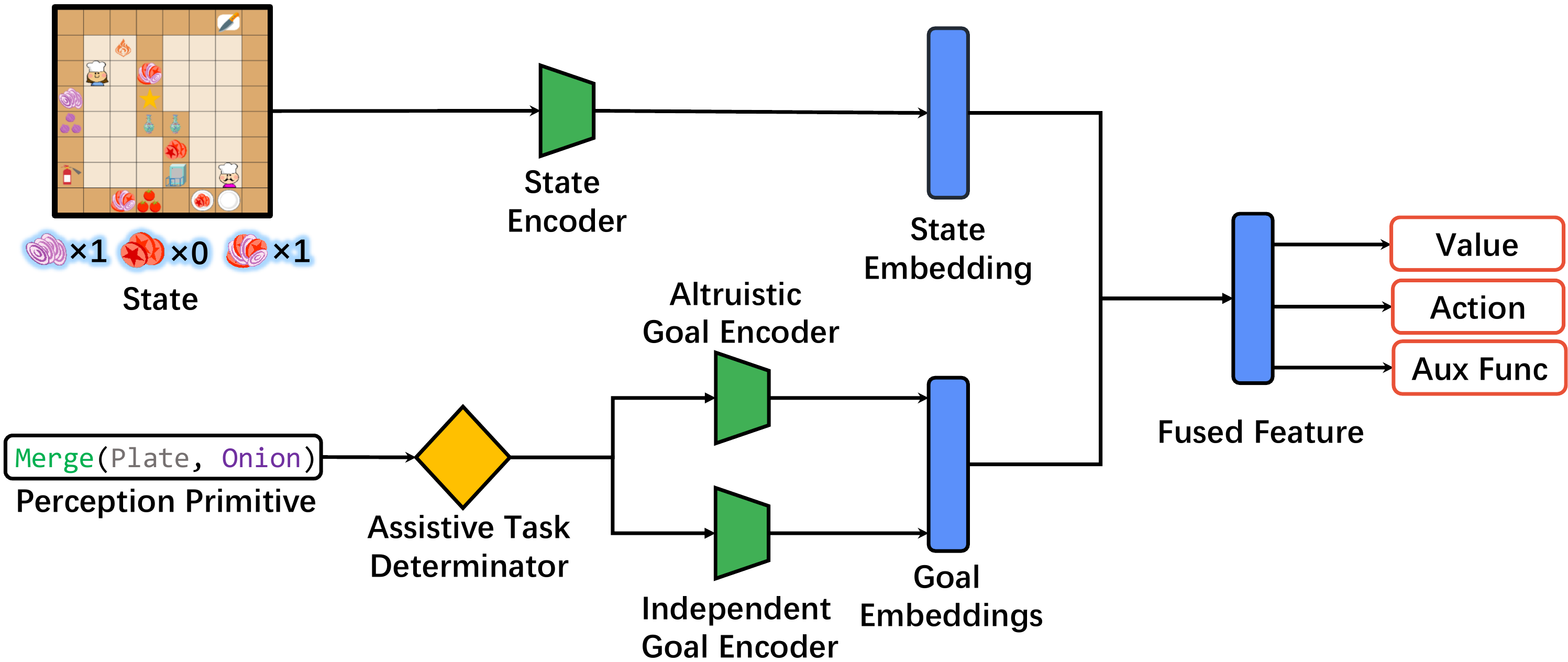}
\caption{\textbf{The policy module}. We adopt a goal-conditioned reinforcement learning framework to determine the policy. 
We encode goals with a goal encoder and fuse them into the state features. If an agent is assistive, the goal is encoded with an altruistic goal encoder instead. The network has three outputs, including the value function, the action distribution, and the auxiliary functions.
}
\label{fig:policy module}
\end{figure}

\subsection{Policy Module}\label{sec:pol_mod}

The policy module grounds agents' actions with the programs and encourages cooperation in completing the tasks.
The overall policy learning procedure advocates a subtask-conditioned reinforcement learning framework as shown in Figure~\ref{fig:policy module}{}. Our algorithm backbone is MAPPO~\citep{yu2021surprising} and consists of two stages. More concretely, we first learn a policy for each agent where other agents' policies are fixed. The input is the state and the encoded subtask. The reward signal is based on whether a subroutine is executed correctly. Then we enable multiple agents to coordinate by learning a joint policy for accomplishing collective goals cooperatively and efficiently, with the help of auxiliary functions and a task allocator.
We use the same architecture to fuse state and goal features as that in the perception module in Figure~\ref{fig:perception module}. We also leverage self-imitation learning to tackle the challenge of sparse rewards, which is shown in Appendix~\ref{app:sil}

\paragraph{Learning to cooperate on a subtask.} 
After obtaining single-agent policies, we then encourage agents to accomplish harder tasks that require cooperation. When assigned a cooperative subtask, one of the agents is in charge of finalizing the task. The rest of the agents are assistive. For example, the assistive agents might pass an onion to the leading agents to chop.
In practice, we randomly appoint one agent as the leading agent and the others as the assistive agents. The leading agent is rewarded if this subtask is completed. The reward function for the assistive agents is calculated based on the \emph{reachability} improvement of the leading agent. We defer the formal definition of this reward function to Section~\ref{sec: reachability}. Such reward function encourages the assistive agents to help the leading agents to obtain rewards. Then, we use a MAPPO-style algorithm to obtain a cooperation policy. We also encode the leading agent's goal into the observation space of the assistive agents to enhance information sharing.

\begin{figure}[t] 
 \centering
     \begin{subfigure}[b]{0.3\textwidth}
         \centering
         \includegraphics[trim=390 165 390 165, width=0.78\textwidth, clip]{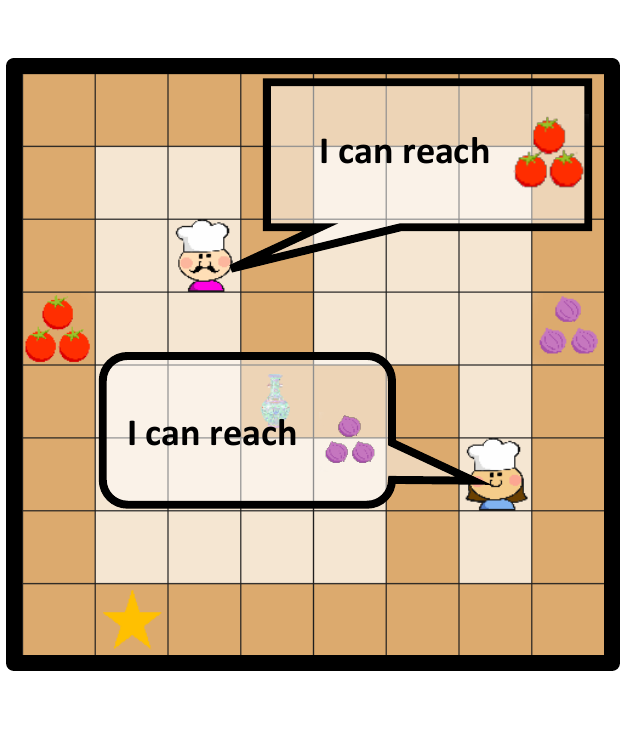}
         \caption{reachability}
         \label{fig:reachability}
     \end{subfigure}
     \hfill
     \begin{subfigure}[b]{0.3\textwidth}
         \centering
         \hspace*{-10.0pt}\includegraphics[trim=360 160 360 160, width=1.0\textwidth, clip]{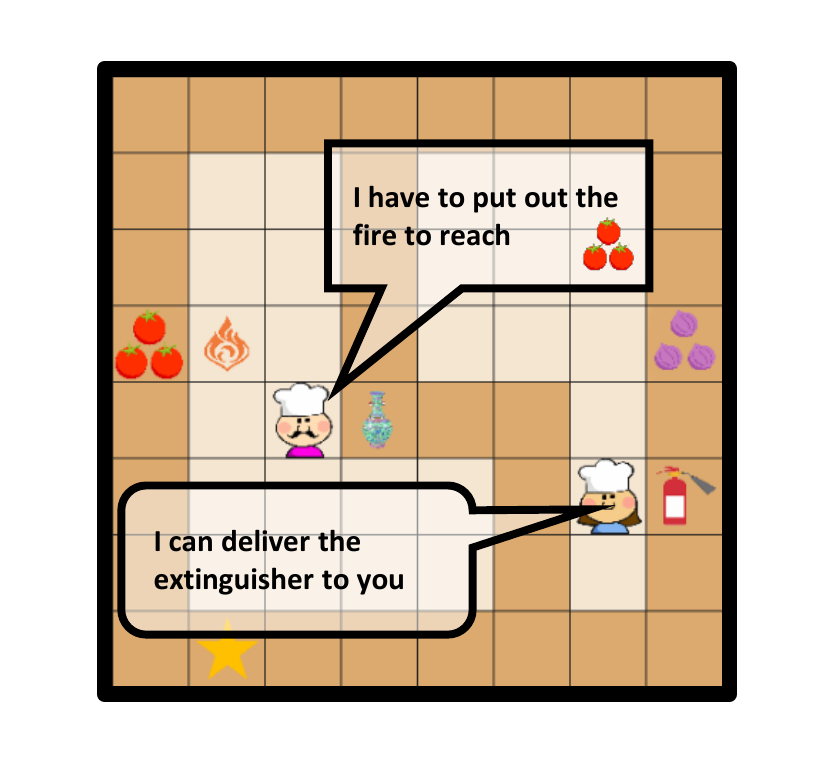}
         \caption{feasibility}
         \label{fig:feasibility}
     \end{subfigure}
     \hfill
     \begin{subfigure}[b]{0.3\textwidth}
         \centering
         \hspace*{-10.0pt}\includegraphics[trim=360 160 360 160, width=1.0\textwidth,  clip]{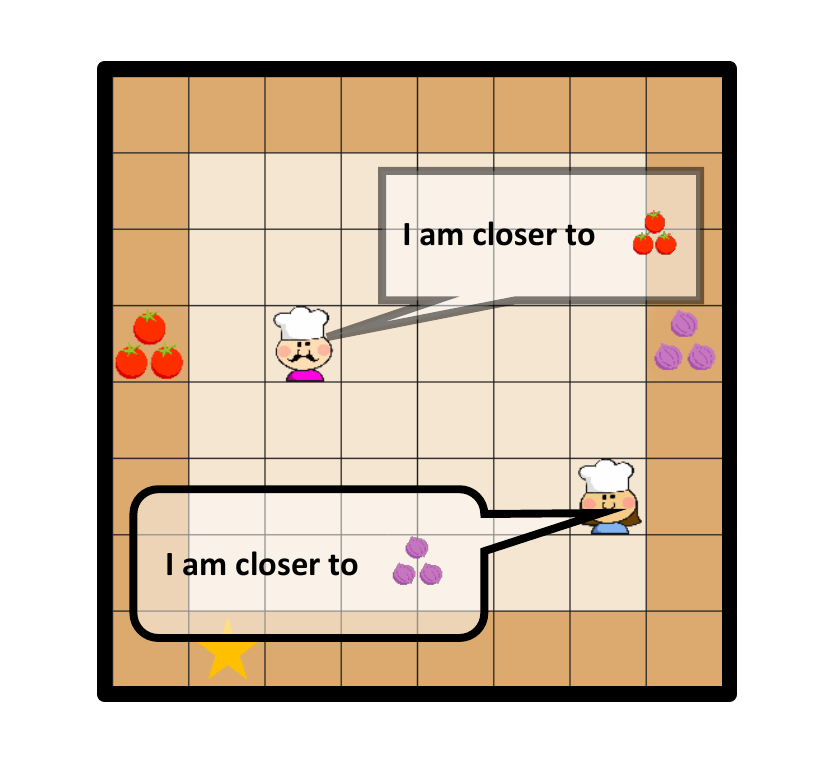}
         \caption{cost-to-go}
         \label{fig:cost-to-go}
     \end{subfigure}
        \caption{\textbf{The auxiliary functions}. Figure~\ref{fig:reachability}{} shows the reachability that indicates if an agent can reach an object. Figure~\ref{fig:feasibility}{} shows the feasibility that indicates whether a subtask can be achieved by the agents collectively. Figure~\ref{fig:cost-to-go}{} shows the cost-to-go function that indicates time consumption.}
        \label{fig:auxiliary functions}
\end{figure}
\subsection{Task Allocator}
The task allocator assigns subtasks to each agent to accomplish a long-horizon task together. Specifically, we design the task allocator for efficient coordination based on the following principles: 1)~It only assigns  subtasks that are feasible for the agent(s) without additional prerequisite subtasks.  2)~If a subtask is cooperative, the task allocator would assign the subtask to a number of agents. 3)~It assigns a subtask to the agent that has the lowest cost in terms of execution time.

To facilitate the task allocation, we propose to learn three auxiliary functions: a \emph{reachability function}, a \emph{feasibility function}, and a \emph{cost\text{-}to\text{-}go function} as illustrated in Figure~\ref{fig:auxiliary functions}{}. Training details of the auxiliary functions are in Appendix~\ref{app:aux}

\label{sec: reachability}
\paragraph{Reachability.} The reachability function $f_{reach}=f_{reach}(s_t,i,\tau)$ is defined as a boolean value to indicate if agent $i$ can complete task $\tau$ alone at the state $s_t$. For a cooperative subtask $\tau$ and a selected leading agent $i$, an extrinsic reward $f_{reach}(s_t^{'},i,\tau)-f_{reach}(s_t,i,\tau)$ is provided to the assistive agents for learning altruistic behaviors.
To train this reachability function, we randomly sample triplets $(s_t,i,\tau)$ and obtain the ground-truth value $f_{reach}^{gt}$ through running the pre-trained non-cooperative policy. Then we optimize the network by minimizing the binary cross entropy $\text{BCE}(f_{reach}^{pred},f_{reach}^{gt})$.   

\paragraph{Feasibility.} The feasibility function $f_{feas}=f_{feas}(s_t,i,\tau)$ is defined as a boolean variable indicating if an agent $i$ can complete a subtask $\tau$ with others' assistance at the state $s_t$. The training procedure of the feasibility function is similar to that of the reachability function, except that we leverage the cooperative policy instead of the single agent policy to collect training data.

\paragraph{Cost-to-go.} The cost-to-go function $f_{cost}=f_{cost\text{-}to\text{-}go}(s_t,i,\tau)$ denotes the remaining timesteps for agent $i$ to accomplish the subtask from the state $s_t$.
We leverage trajectories generated by a pre-trained intermediate cooperative policy from \model to train this cost function. Specifically, we randomly sample triplets $(s_t,i,\tau)$ and execute the pre-trained cooperative policy in Section ~\ref{sec:pol_mod} to obtain the ground-truth timesteps $f_{cost\text{-}to\text{-}go}^{gt}$ to complete the subtask $\tau$. 
Then we optimize the network by reducing the Mean Squared Error~(MSE) $\mathcal{L}_{cost\text{-}to\text{-}go}=\text{MSE}(f_{cost\text{-}to\text{-}go}^{pred},f_{cost\text{-}to\text{-}go}^{gt})$.

\paragraph{Criteria for subtask allocation.} For a specific Possible Subroutine Set $\{\tau_1,\dots,\tau_m\}$, we denote a legal subtask allocation as $\{T_1,T_2,\dots,T_m\}$ such that \begin{itemize}
    \item $T_i$ is a list of $n_i$ agents where $n_i=|T_i|$.  The agents are those in the agent set $\{1,\dots,N\}$ who aim at completing the subtask $\tau_i$.
    \item For all $i\in \{1,\dots,m\}$,  if $n_i$ is larger than one, then the agent $T_i^1$ is selected as the leading agent and the agent(s) $T_i^2,\dots,T_i^{n_i}$ are selected as the assistive agent(s).
    \item For $1 \leq i < j \leq m$, we have that $T_i\cap T_j=\emptyset$, 
    which means no agent is assigned two subtasks simultaneously.
\end{itemize} 
We compute the cost of each possible subtask allocation $\{T_1,T_2,\dots,T_m\}$ as the sum of three terms
\begin{align}
    c_{\text{total}}=c_{\text{feas}}+c_{\text{cost-to-go}}+c_{\text{reach}}
\end{align},where
\begin{align}
  c_{\text{feas}}= & \sum_{i=1}^{m} -n_i w_{\text{feas}} \log f_{\text{feas}}(s,T_i[1],\tau_i)
\end{align}
\vspace{-0.8cm}
\begin{align}
  c_{\text{cost-to-go}}= & \sum_{i=1}^{m} -n_i w_{\text{cost}}  f_{\text{cost}}(s,T_i[1],\tau_i)
\end{align}
\vspace{-0.8cm}
\begin{align}
    c_{\text{reach}}=\sum_{i=1}^{m} \mathbb{I}[n_i=1] w_{\text{reach}} \log f_{\text{reach}}(s,T_i[1],\tau_i)
\end{align}
$w_{feas}$, $w_{reach}$, and $w_{cost}$ are tunable hyperparameters and $\mathbb{I}[\cdot]$ is the indicator function. The \emph{feasibility} term encourages the agents to choose the feasible subtasks, the \emph{cost-to-go} term encourages the agents to choose the less costly subtasks, and the \emph{reachability} term guarantees that the subtask assigned to only one agent is non-cooperative. The $log$ operator operating on a value approximating $1$ will induce a huge cost, thus preventing the allocation of an infeasible or unreachable subtask to an agent. We search from the possible allocations $\{T_1,T_2,\dots,T_m\}$ and apply the one with minimal cost as the final allocation.

In practice, we also use two hyperparameters: $c_r$ to be subtracted from the cost-to-go function to encourage the agents to finish their ongoing subtasks, and $c_i$ to be added to the cost-to-go function to avoid allocating subtasks to an agent who can never accomplish a task within a timeout threshold $t_o$.

\subsection{Complexity Analysis}
In an environment with $M$ subtasks and $N$ agents, the brute force search for an optimal allocation indeed has a complexity of $O(M^N)$. However, the practical complexity is much smaller than it. The reasons are as follows:
\begin{enumerate}
    \item In a certain stage of a long-horizon task, only a small amount of subtasks are feasible. Thus,  the subtask amount $M$ can be pruned into a smaller number $L$ by checking the feasibility function $O(M\times N)$ times and ignoring the subtasks whose feasibility functions are less than a given threshold. 
    \item The engaging $N$ agents can be classified into $C$ roles. The agents sharing the same role have the same reachability functions. $C$ is often a property of the task that does not scale with $N$. For example, in the overcooked environment, $C$ can be the number of connected components of the map. Note that, in E-MAPP, the assistive agents aim to increase the reachability of the leading agents. We define $C\times L$ new subtasks $(\tau, c)$, where $\tau$ comes from the $L$ feasible subtasks and $c$ comes from the $C$ roles. The goal of each new subtask $(\tau, c)$ is to help agents with role $c$ to gain reachability on subtask $\tau$. We can obtain a new subtask set of size $O(C\times L)$ by extending the original subtask set with these newly defined subtasks. Assume that the number of agents is smaller than the number of feasible subtasks (otherwise, idle agents will inevitably emerge). Under this assumption, each agent will choose to either complete a subtask alone or assist a certain group of agents with the same role, and each subtask in the new subtask set is allocated to at most one agent to avoid conflict. Then the task allocation problem turns into finding the best matching of $N$ agents and $O(C\times L)$ subtasks with the smallest total cost, which can be solved by the Hungarian algorithm. The  computational complexity is $O((N+CL)^3)\leq O((N+CM)^3)$ that scales well.
\end{enumerate}

\section{Experiments}
In this section, we aim to investigate the following key questions. First, is the program guidance helpful for agents to understand long-horizon tasks in comparison with other structured information guidance? Second, are the parallel structures in the programs bring forth cooperative behaviors among the agents? Third, does the task allocator improve the completion rates and the time efficiency of the long-horizon tasks by virtue of the auxiliary functions? 

\subsection{Environment Description}

\begin{wrapfigure}{r}{0.30\textwidth} 
\vspace{-.2in}
\centering
\includegraphics[width=0.30\textwidth]{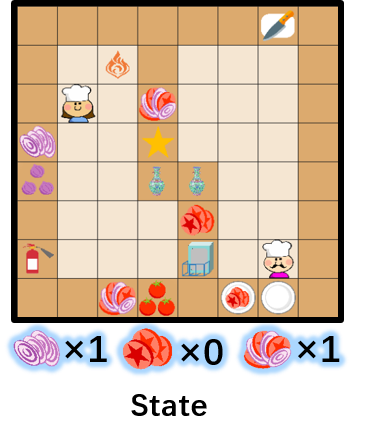}
\caption{A sample from the \textit{Overcooked} environment.}
\vspace{-.4in}
\label{fig:overcooked_demo}
\end{wrapfigure}
To evaluate the proposed framework, we adapted the previous environment~\cite{wu2021too}mimicking the video game to a more challenging one ``Overcooked v2''. Concretely, we extend temporally for the horizon length of a task by adding additional behaviors such as ``wash dishes'' and ``put out fire''. As shown in Figure~\ref{fig:overcooked_demo}, agents can navigate through the grid world, interact with objects~(e.g., tomatoes, knives), and deliver the dishes to the customers~(the yellow star). The goal of the agents is to serve dishes according to the given recipes that can be divided into subtasks. More details about the environment can be found in Appendix~\ref{app:env}.

\subsection{Setup}
\noindent \textbf{Tasks.}
We test the agents on a variety of tasks with different difficulty levels. The {\bf easy} tasks contain only one subroutine for verification. The {\bf medium} tasks contain two or three subroutines. The {\bf hard} task requires the agents to complete multiple dishes that require at least two subroutines each while putting out the randomly appearing fire. We conduct evaluation in two different patterns, based on seen or unseen tasks. Note that both evaluation scheme are conducted on novel maps, which makes the task more challenging. The unseen tasks share the subroutines with the seen tasks but is never used as a goal during training. The unseen tasks are used to test whether the methods can generalize compositionally.  We also note that for the tasks with a \textit{repeat} command~(e.g., repeatedly pick an onion from the supply), a targeted repeat number is preset.

\noindent \textbf{Metrics.}
We use the completion rates and the average scores as the metrics. The completion rate is defined as the percentage of tasks that can be completed in an episode. The average score is the discounted cumulative rewards in an episode across all testing tasks. Agents will receive a reward of $1$ when the final goal is achieved and receive a reward of $0.2$ when a correct subtask is completed. Each algorithm is tested in 1000 environments and we report their average in the tables.

\noindent \textbf{Baselines.} We compare our model with two baselines: MAPPO~\cite{yu2021surprising} and a natural language-guided agent. Detailed descriptions are in Appendix~\ref{app:baseline}.

\subsection{Results}
\paragraph{Results with seen tasks in novel maps.}
Table~\ref{tab:seen task}{} shows the completion rate and average scores on seen tasks in novel maps, demonstrating that E-MAPP excels at understanding the structure of complex tasks. As expected, the end-to-end MAPPO model performs well on short-horizon tasks, but suffers from a significant performance drop when the task becomes complex. The natural language--guided model also has a large performance drop when the horizon of the tasks becomes longer. We attribute this to the agents' failure to understand the complex task structure described in natural language without explicit structure. By contrast, \model performs well even when the tasks have a long horizon with various accidental events happening.  

\paragraph{Generalization with unseen tasks in novel maps.}
Table~\ref{tab:unseen task} shows the results on the zero-shot generalization scheme. The scenarios and tasks are very different from the training domain, posing an extra challenge to all the algorithms, including the auxiliary functions of \model. We find that \model is significantly better than both the previous multi-agent RL algorithms and a language-guided agent.  We attribute the success of \model to the fact that programs have better compositionality and less ambiguity. In the training time, \model learns certain subroutines from other tasks; it applies the behavior compositionally to achieve goals never seen during training. More visualized results are in Appendix~\ref{app:gen}.

\begin{table}[t]
\setlength{\tabcolsep}{4pt}

  \caption{\textbf{Results on the seen tasks in novel maps.} Completion rates and average scores of seen tasks in the \textit{Overcooked} environment with three difficulty levels.}
  \label{tab:seen task}
  \centering
\begin{tabular}{lcccccc}
\toprule[0.5mm]
& \multicolumn{3}{c}{Completion Rates}                  & \multicolumn{3}{c}{Average Scores} \\ 
\cmidrule(lr){2-4}\cmidrule(lr){5-7}
Methods & Easy  & Medium & Hard & Easy  & Medium & Hard \\ \midrule
\model (ours)   &  98.0\%& $\mathbf{97.5\%}$ & $\mathbf{56.3\%}$ & 1.06$\pm$0.17 &  \textbf{1.12}$\pm$0.22    &  \textbf{1.58}$\pm$0.60\\
Natural language guidance & 81.9\% & 48.1\% & 1.01\% & 0.87$\pm$0.43 & 0.63$\pm$0.51  &  0.82 $\pm$ 0.31\\
MAPPO~\cite{yu2021surprising}   & $\mathbf{100.0\%}$ & 65.7\%  & 0.00\% & \textbf{1.11}$\pm$0.04 &0.79$\pm$0.31  &  0.59$\pm$ 0.27 \\
\bottomrule
\end{tabular}
\end{table}

\begin{table}[ht]
  \caption{\textbf{Results on the unseen tasks in novel maps.} Completion rates and average scores of unseen tasks with two difficulty levels.}
  \label{tab:unseen task}
  \centering
\begin{tabular}{lcccc}
\toprule[0.5mm]
      & \multicolumn{2}{c}{Completion Rates}                  & \multicolumn{2}{c}{Average Scores} \\ 
\cmidrule(lr){2-3}\cmidrule(lr){4-5}
 Methods  & Medium & Hard  & Medium & Hard \\ 
 \midrule
E-MAPP (ours)    & \textbf{100.0\%} & \textbf{43.7\%} & \textbf{1.13} $\pm$ 0.10 & \textbf{0.99} $\pm$ 0.22 \\ 
natural language--guided model  & 58.4\% & 0.0\%  &  0.58 $\pm$ 0.50  &  0.48 $\pm$ 0.21\\
\bottomrule
\end{tabular}
\end{table}

\paragraph{Ablation of auxiliary functions.} We analyze the significance of the proposed key components in our model by comparing our model with the variants that removes~1)~the feasibility predictor~2)~the reachability predictor~3)~the cost-to-go predictor. The completion rates and the average scores are shown in Table~\ref{tab:ablation}. We find that all the components are important to \model. Specifically, we find that removing the reachability predictor from \model leads to a significantly increased average timestep, and this is the component that provides the largest performance gain. We attribute this to the reachability function that pointed out the subtasks that can be completed individually, providing an impetus for agents to parallelize.

\paragraph{Ablation of parallelism in the programs.} We compare the proposed parallel programs with sequential programs in the same environments. In Table~\ref{tab:ablation}, we find that the agents with parallel programs use 15\% fewer time steps to accomplish the same goal as those with sequential programs.  These results show that the parallel programs are central to \model in terms of time efficiency.

\begin{table}[ht]
  \caption{\textbf{Ablation study.} Completion rates, average scores, and timesteps when one of the key components of \model is removed. }
  \label{tab:ablation}
  \centering
\begin{tabular}{lcccccc}
\toprule[0.5mm]
     & Completion Rates  $\uparrow$              & Average Scores $\uparrow$ & Average Timesteps $\downarrow$ \\ \midrule 
\model & \textbf{56.3\%} & \textbf{1.58} $\pm$ 0.60 & \textbf{17.6}&  \\ \midrule
w/o feasibility predictor   & 38.5\%   & 1.42 $\pm$ 0.56 & 21.24 \\   
w/o reachability predictor & 43.8\% & 1.45 $\pm$ 0.50 &23.37 \\
w/o cost-to-go predictor  & 52.0\% & 1.46 $\pm$ 0.45  & 20.42 \\
sequential program & 48.8\% & 1.51 $\pm$ 0.47 & 20.33 &  \\ \bottomrule
\end{tabular}
\end{table}

\paragraph{Partially observable environments.} We conduct an experiment to show that E-MAPP can still outperform other methods in a partially observable environment. In the new setting, the observation of each agent is only part of the map within reach. Table ~\ref{tab:partial obs} shows the results. Under the new setting, E-MAPP can still learn to allocate sub-tasks to agents and accomplish tasks efficiently. 
\begin{table}[ht]
\caption{\textbf{Additional experiment in partially observable environments} we conduct an experiment to show that E-MAPP can still outperform other methods in a partially observable environment.}
  \label{tab:partial obs}
  \centering
  \begin{tabular}{ccc}
   \toprule[0.5mm]
   model & score & completion rate\\  \midrule
   E-MAPP (partial obs) & 1.01±0.38 & 27.1\%\\
   E-MAPP (original) & 1.58±0.60  & 56.3\% \\
   MAPPO &  0.59± 0.27 & 0.0\% \\
\bottomrule
\end{tabular}
\end{table}

\subsection{Visualization of Learned Behaviors}
\begin{figure}[ht] 
 \centering
     \begin{subfigure}[b]{0.3\textwidth}
         \centering
         \includegraphics[width=\textwidth]{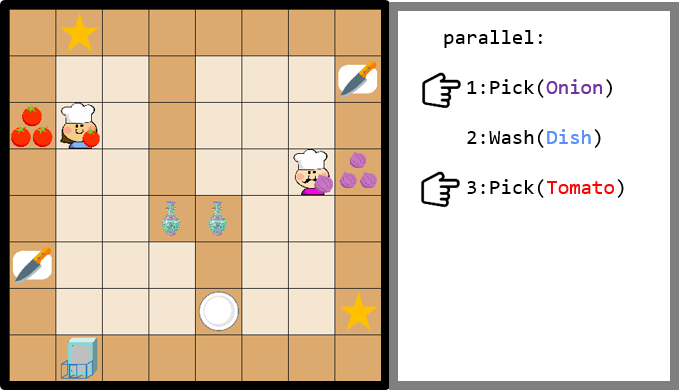}
         \caption{parallel}
         \label{fig:parallel}
     \end{subfigure}
     \hfill
     \begin{subfigure}[b]{0.3\textwidth}
         \centering
         \includegraphics[width=\textwidth]{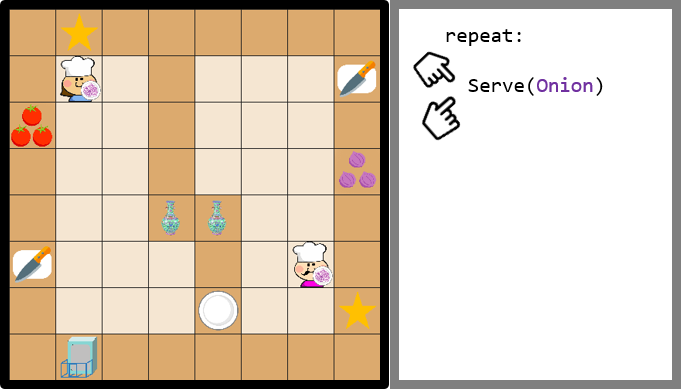}
         \caption{repeat}
         \label{fig:repeat}
     \end{subfigure}
     \hfill
     \begin{subfigure}[b]{0.3\textwidth}
         \centering
         \includegraphics[width=\textwidth]{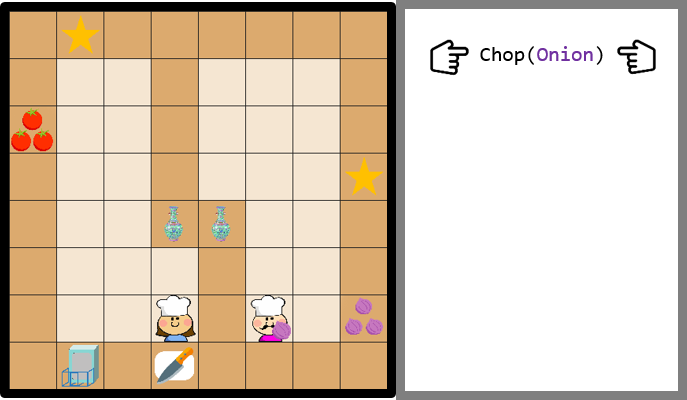}
         \caption{assist}
         \label{fig:assist}
     \end{subfigure}
        \caption{\textbf{Learned behaviors of \model}. Figure~\ref{fig:parallel} shows that two agents are assigned two subtasks concurrently. Figure~\ref{fig:repeat} shows that both agents are trying to repeatedly perform the same task. Figure~\ref{fig:assist} shows that one agent is passing an onion to the agent who can chop it.}
        \label{fig:viz}
\end{figure}
In Figure~\ref{fig:viz}, we show the representative learned behaviors of \model. In Figure~\ref{fig:viz}(a), the two subroutines are allocated to two agents separately. This demonstrates the effectiveness of the \textit{feasibility} function, with which the task allocator successfully identified the parallelizable subtasks that can be executed concurrently. We note that the \textit{parallel} indicator in the program only suggests the agent trying to identify what subtasks are parallelizable rather than providing a ground-truth task structure. In Figure~\ref{fig:viz}(b), we find that the two agents are performing the same task in a \textit{repeat} subroutine independently. We show the allocator can assign tasks to multiple agents that are reachable to the goal. In Figure~\ref{fig:viz}(c), we show that an assistive agent is passing an onion to the leading agent to chop. This shows how the agents learn to cooperate with each other. More visualizations can be found in Appendix~\ref{app:gen}.

\subsection{Scalability to New Domains} We also investigate whether \model can be effective on tasks with continuous action space based on the \emph{Stacking} environment. More details and demos in \emph{Stacking} can be found in Appendix~\ref{app:robot}. We discuss some potential parallel-program synthesis approaches when applying E-MAPP to a new domain in Appendix~\ref{app:program synthesis}

\section{Conclusion}
In this paper, we first formulate the problem of program-guided multi-agent tasks. We propose Efficient Multi-Agent Reinforcement Learning with Parallel Programs~(E-MAPP), an effective framework that uses a type of parallelism-aware program for multi-agent collaboration and a task allocation strategy via a set of learnable auxiliary functions. The results show that our algorithm can infer the task structure and significantly boost completion rates, efficiency, and generalization ability on long-horizon tasks.

\paragraph{Limitations.} Our current framework does not consider more challenging scenarios such as dynamic scenes or generalization to novel objects. We believe that this work opens a welcoming avenue to this research direction, and more future works will address additional challenges.

\paragraph{Acknowledgements.}
We thank Yuping Luo and Zhecheng Yuan for their careful proofreading and writing suggestions.

\clearpage
\bibliographystyle{plain}
\bibliography{reference}
\clearpage
\section*{Checklist}

\begin{enumerate}

\item For all authors...
\begin{enumerate}
  \item Do the main claims made in the abstract and introduction accurately reflect the paper's contributions and scope?
    \answerYes{}
  \item Did you describe the limitations of your work?
    \answerYes{}
  \item Did you discuss any potential negative societal impacts of your work?
    \answerNA{Our work focus on designing an algorithm for boosting the performance and generalization ability, rather than a specific real-world application. }
  \item Have you read the ethics review guidelines and ensured that your paper conforms to them?
    \answerYes{}
\end{enumerate}

\item If you are including theoretical results...
\begin{enumerate}
  \item Did you state the full set of assumptions of all theoretical results?
    \answerNA{}
        \item Did you include complete proofs of all theoretical results?
    \answerNA{}
\end{enumerate}

\item If you ran experiments...
\begin{enumerate}
  \item Did you include the code, data, and instructions needed to reproduce the main experimental results (either in the supplemental material or as a URL)?
    \answerYes{In the supplementary material}
  \item Did you specify all the training details (e.g., data splits, hyperparameters, how they were chosen)?
    \answerYes{In the supplementary material}
        \item Did you report error bars (e.g., with respect to the random seed after running experiments multiple times)?
    \answerYes{}
        \item Did you include the total amount of compute and the type of resources used (e.g., type of GPUs, internal cluster, or cloud provider)?
    \answerYes{In the supplementary material}
\end{enumerate}

\item If you are using existing assets (e.g., code, data, models) or curating/releasing new assets...
\begin{enumerate}
  \item If your work uses existing assets, did you cite the creators?
    \answerYes
  \item Did you mention the license of the assets?
    \answerYes
  \item Did you include any new assets either in the supplemental material or as a URL?
    \answerYes
  \item Did you discuss whether and how consent was obtained from people whose data you're using/curating?
    \answerNA{}
  \item Did you discuss whether the data you are using/curating contains personally identifiable information or offensive content?
    \answerNA{}
\end{enumerate}

\item If you used crowdsourcing or conducted research with human subjects...
\begin{enumerate}
  \item Did you include the full text of instructions given to participants and screenshots, if applicable?
    \answerNA{}
  \item Did you describe any potential participant risks, with links to Institutional Review Board (IRB) approvals, if applicable?
    \answerNA{}
  \item Did you include the estimated hourly wage paid to participants and the total amount spent on participant compensation?
    \answerNA{}
\end{enumerate}

\end{enumerate}

\clearpage

\appendix

\section{Appendix}
The project website can be viewed at \href{https://sites.google.com/view/e-mapp}{https://sites.google.com/view/e-mapp}.
\subsection{Domain Specific Language~(DSL) Specifications}
\label{app:dsl}
Table~\ref{tab:dsl} shows the domain-specific language~(DSL) designed for \model in the \textit{Overcooked-v2} environment. 

\begin{table}[ht]
    \centering
    \caption{\textbf{DSL for the Overcooked-v2 Environment.} We list types of constituent elements of the DSL. We provide the context-free grammar for generating all legal programs.}
    \begin{tabularx}{\textwidth}{lX} \toprule
        Type & Instances\\ \midrule
        Program {\tt p} & {\tt def main():s} \\ \midrule
        Item \tt{t} & {\tt FreshOnion} $\mid$ {\tt FreshTomato} $\mid$ {\tt Plate} $\mid$ {\tt ChoppedOnion} $\mid$ {\tt ChoppedTomato} $\mid$ {\tt ChoppedOnion+Plate} $\mid$ {\tt ChoppedTomato+Plate} $\mid$ {\tt ChoppedOnion+ChoppedTomato} $\mid$ {\tt ChoppedOnion+ChoppedTomato+Plate} \\ \midrule
        Behavior {\tt b} & {\tt Chop(t)} $\mid$ {\tt Pick(t)} $\mid$ {\tt Merge(t,t)} $\mid$ {\tt Serve(t)} $\mid$ {\tt WashDirtyPlate()} {\tt PutOutFire()}\\ 
        \midrule
        Conditions {\tt c} & {\tt h} $\mid$ {\tt tautology} \\ \midrule
        Statement {\tt s} & {\tt While(c):(s)} $\mid$ {\tt Parallel:(s\_1, s\_2, $\cdots$)} $\mid$ {\tt b} $\mid$ {\tt Repeat(s,i)} $\mid$ {\tt If(c):(s)} $\mid$  {\tt else:(s)} $\mid$ \\ \midrule
        Perception {\tt h} & {\tt is\_ordered[t]} $\mid$  {\tt is\_there[t]} \\
        \bottomrule
    \end{tabularx}
    \label{tab:dsl}
\end{table}

\subsection{Multi-Pointer Updater}
\label{app:mpu}
Algorithm~\ref{alg: multi-pointer update} shows the execution rules of parallel programs. The program executor maintains a set of pointers $\sP$ pointing to different subroutines. After the agents take the actions, we update each pointer $p$ in $\sP$ according to the type of the subroutine that $p$ was pointing to.
\begin{algorithm}
\scriptsize
\SetKwData{Left}{left}\SetKwData{This}{this}\SetKwData{Up}{up} \SetKwFunction{Union}{Union}\SetKwFunction{FindCompress}{FindCompress} \SetKwInOut{Input}{input} \SetKwInOut{Output}{output}
\caption{Multi-Pointer Updater}\label{alg: multi-pointer update}

\Input{a set of pointers $\sP$, a required repetition number $M$} 
\For{each pointer $p$ in $\sP$}{

\If{$p$ points to an \emph{if}-routine}{Point $p$ to the routine inside/outside the block of primitives in the if-routine if the response to its condition  is True/False.}
\ElseIf{p points to an \emph{while}-routine}{Point $p$ to the routine inside/outside the block of primitives in the while-routine if the response to its condition  is True/False.}
\ElseIf{$p$ points to a \emph{parallel}-routine}{Split $p$ into multiple pointers, each pointing to a block of subroutines in the parallel-routine. }
\ElseIf{$p$ points to a \emph{repeat}-routine}{Split $p$ into M pointers, each pointing to a copy of the block of subroutines.}
\ElseIf{$p$ points to a behavior primitive and the corresponding subtask is completed}
{
\If{$p$ points to the end of a block of subroutines in a \emph{while}-routine}{Point $p$ to the while-routine.}
\ElseIf{$p$ points to the end of a block of subroutines in a group of blocks of subroutines spawned from a \emph{parallel}-routine/\emph{repeat}-routine}
{
\If{the pointed block is not the last remaining one}{remove the pointer}
\Else{remove the pointer and generate a new pointer $q$ pointing to the subroutine subsequent to the \emph{parallel}-routine/\emph{repeat}-routine}
}
\Else{Point $p$ to the subsequent subroutine. Terminate the program if no subsequent subroutine exists.}
}
}
\end{algorithm}

\subsection{Full Illustration of the \model Algorithm}
\label{app:algo}
We describe the whole process of  \model in Algorithm~\ref{alg:full algorithm}, which corresponds to the inference stage of E-MAPP. 
\begin{algorithm} \SetKwData{Left}{left}\SetKwData{This}{this}\SetKwData{Up}{up} \SetKwFunction{Union}{Union}\SetKwFunction{FindCompress}{FindCompress} \SetKwInOut{Input}{input} \SetKwInOut{Output}{output}
\caption{The E-MAPP Algorithm during test time}\label{alg:full algorithm}
\Input{Environment $env$ and its guiding program, a set of pointers $\sP$ pointing to the possible subroutine set, a policy module $f_{policy}$, a perception module $f_{perception}$} 
\While{the task is not completed}{
    \While{there are pending perception primitives in the possible subroutine set}{
    Run $f_{perception}$ on the perception primitives and get the results $o_{perc}$. Update $P$ according to $o_{perc}$ and the rule in Algorithm~\ref{alg: multi-pointer update}.
    }
    {\textbf{Run} $f_{policy}$ to obtain the auxiliary functions on each behavior primitive $p$ points to.}\\
    {\textbf{Compute} the cost of each possible allocation based on the auxiliary functions.}\\
    {\textbf{Run} $f_{policy}$ on the subtasks with minimal cost and obtain the joint action $a$}\\
    {\textbf{Environment} steps forward with $a$}\\
    \While{There are completed behavior primitives in the possible subroutine set}{Update $p$ according to the rule in Algorithm~\ref{alg: multi-pointer update}
    }
}
\end{algorithm}

\subsection{Training Details for the Policy Module}
\paragraph{Architecture details.}
\label{policy arch}
The policy module takes as input a goal vector $g$ and a state $s$  and outputs an action distribution $\tilde{a}$. A goal vector $g$ has a size of $20$ where the first $10$ elements are one-hot encoded behavior type~(e.g., $\emph{Chop}$) and the latter $10$ elements are one-hot encoded behavior arguments~(e.g., $\emph{Tomato}$). A state $s$ is comprised of a map state $s_{\text{map}}$ and an inventory state $s_{\text{inv}}$. The map state $s_{\text{map}}$ has a size of $20\times 8\times 8$ where $8\times 8$ is the resolution of the map and $20$ is the number of object types. The inventory state $s_{\text{inv}}$ has a size of $6$ where the first two entries denote the location of the agent, the third entry denotes whether the agent is holding objects, and the last three entries denote whether a certain dish is ordered.

The map state $s_{\text{map}}$ is encoded by a four-layer convolutional neural network~(CNN) with channel sizes of 32,64,64, and 64. For each convolutional layer, we use $1$ as the stride size and $1$ as the ``same" padding size. Each convolutional layer has a kernel size of $3$ except for the first one, which has a kernel size of $5$. A ReLU nonlinearity is applied to every convolutional layer. The output is finally flattened into a feature vector and fed into a linear layer, producing a $128$-dim map feature vector, denoted as $f_{\text{map}}$.

The inventory state $s_{\text{inv}}$ is encoded by a three-layer MLP with hidden size 128 for all layers. The output feature vector has a dimension of $32$, denoted as $f_{\text{inv}}$.

The map feature $f_{\text{map}}$ and the inventory feature $f_{\text{inv}}$ are then concatenated, producing the $160$-dim state feature $f_{\text{state}}^1$. 

The goal vector $g$ is encoded by a three-layer MLP with a hidden size of 128 for all the layers. The output goal feature $f_{goal}$ is a $640$-dim feature vector. $f_{goal}$ can be viewed as the concatenation of four
$160$-dim feature vectors,denoted as $\gamma_1$, $\beta_1$, $\gamma_2$, and $\gamma_2$ respectively.

The state feature is modulated by $\gamma_1$ and $\beta_1$ as $f_{\text{state}}^2=f_{\text{state}}^1\dot \gamma_1+\beta_1$. Next, the modulated feature $f_{\text{state}}^2$ is encoded by a two-layer MLP with both hidden size and output size equal to 128. The output is then modulated by $\gamma_2$ and $\beta_2$, producing the goal-conditioned state feature $f_{\text{state}}$. 

Finally, the goal-conditioned state feature $f_{\text{state}}$ is encoded by two linear layers to produce the $24$-dim action distribution and the $1$-dim value function.

In practice, the parameters of the convolutional layers are shared by the policy net and the value net, while other parameters are separate.

In cooperative settings, the goal input of the assistive agent is the leading agent's goal. We use a separate assistive goal encoder for it. The architecture of the altruistic goal encoder is the same as the independent goal encoder mentioned above.

\paragraph{Self-imitation learning.}
In our task domain, it is important to address the challenge of sparse rewards, which is also a key issue for goal-conditioned reinforcement learning~\cite{pmlr-v80-oh18b}. To tackle this, we propose to better utilize the successful trajectories of each agent inspired by self-imitation~\cite{oh2018self}. For each agent, we select the state-action pairs from the replay buffer with empirical returns that are larger than a threshold $r_{\text{thresh}}$. The self-imitation learning objective of the agent $i$ is~$\mathcal{L}_{sil}=-r_i(s,t)\log \pi_i(s_t)$. This loss is added directly to the reinforcement learning loss.
\label{app:sil}

\paragraph{Hyperparameters.} Table ~\ref{tab:policy hyper} shows the hyperparameters used in the policy module. 
\begin{table}[t]
  \caption{\textbf{Hyperparameters for the training of the policy module.} We use MAPPO as the backbone algorithm. The common hyperparameters are listed below.}
  \label{tab:policy hyper}
  \centering
\begin{tabular}{cc}
\toprule[0.5mm]
     Name & Value  \\ \midrule 
     learning rate & 3e-4 \\
     training steps & 10M \\
     update batch size & 256\\
     number of rollout threads & 8\\
     rollout buffer size & 4096$\times$ 8\\
     weight of value loss & 0.1\\
     weight of policy loss & 1 \\
     weight of entropy loss &0.01\\
     \bottomrule
\end{tabular}
\end{table}

\subsection{Architecture and Training Details for the Perception Module.}
\label{app:architecture}
\paragraph{Architecture of the perception module.}
The architecture of the perception module can be obtained from that of the policy module in Appendix~\ref{policy arch} by replacing the goal encoders with perceptive query encoders. The details of the encoders remain unchanged.

\paragraph{Dataset collection.}
We randomly sample $10$k environments as the training dataset. To augment the data, we label the ground truth of the perceptive queries after the agents randomly take $10\text{-}15$ steps. We leave 10\% of the dataset as the evaluation dataset and terminate the training process when the accuracy on the evaluation dataset is larger than 99\% five times.  

\paragraph{Hyperparameters.} Table ~\ref{tab:perception hyper} shows the hyperparameters used in training the perception module. 
\begin{table}[t]
  \caption{\textbf{Hyperparameters for the training of the perception module.} We model the perception module as a binary classifier and use cross-entropy loss as the training objective.}
  \label{tab:perception hyper}
  \centering
\begin{tabular}{cc}
\toprule[0.5mm]
     Name & Value  \\ \midrule 
     learning rate & 3e-4 \\
     update batch size & 128\\
     \bottomrule
\end{tabular}
\end{table}

\subsection{Training Details for the Auxiliary Functions}
\label{app:aux}
We randomly sample $10$k environments as the training environments for each of the three auxiliary functions: $f_{\text{reach}}$, $f_{\text{feas}}$, and $f_{cost\text{-}to\text{-}go}$.

To train the reachability function, we run the pre-trained single-agent policy on the training environments to collect the training data. The agent is required to fulfill a specific subtask within an episode of 128 timesteps. If the agent violates the program by completing the wrong subtask or exceeding the time limit, we will label every state in the trajectory as $False$. In contrast, if the agent successfully completes the subtask, we will label the states as $True$. We train the reachability function by alternatively collecting (state, goal, label) triplets and training on the collected data.

To train the feasibility function, we run the pre-trained multi-agent cooperative policy in the training environments to collect the training data. The agents are required to complete a specific subtask within an episode of 128 timesteps. Although both agents are responsible for the subtask, we only take account of the leading agent's trajectories. The states in the trajectories are labeled True/False according to whether the subtask is successfully completed during an episode. 

To alleviate the problem of mistakenly labeling a feasible (state, goal) pair as $\emph{False}$ due to the imperfection of the pre-trained policy, we run multiple times on the environment that is labeled $\emph{False}$ and correct the label if there is a successful case.

\subsection{Detailed Environment Description}
\label{app:env}
In the $\emph{Overcooked}$ environment, the goal of the agents is to complete long-horizon tasks, such as preparing a dish. A typical dish would require first picking up ingredients from certain supplies, then processing the ingredients (e.g., chop, merge together, or put on a plate), and finally delivering the dish~(and washing the plates if another dish is still needed). We also introduce ``on-fire'' as an accident for the agents to handle with a fire extinguisher. The subtasks correspond to the subroutines in the domain-specific language~(DSL). Different subroutines, together with the control flow, form the guiding parallel programs for the agents.

For each agent, the state is composed of a map state and an inventory state. Both the maps and the programs are one-hot encoded as object-centric representations. The agents' action space is discrete. There are 24 possible actions, including 6 operations~(move, pick, place, serve, merge, and interact) in 4 directions. In multi-agent settings, we follow the original game and assume full observability. 

\subsection{More Visualizations for Generalization}
\label{app:gen}
\begin{figure}[htbp]
	\centering
	\includegraphics[trim=0.3cm 1cm 0.5cm 0.2cm, width=\textwidth]{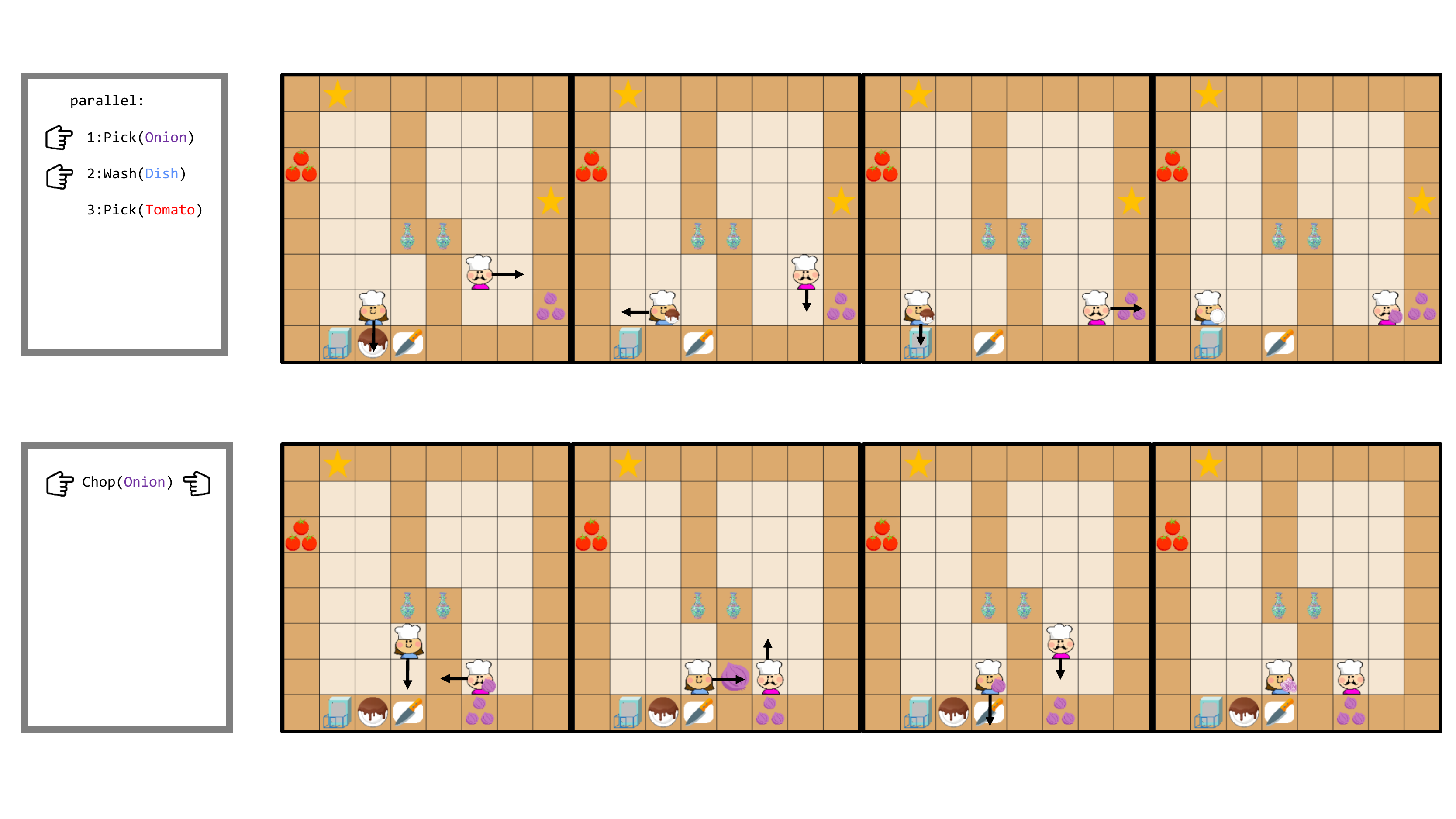}
	\caption{\textbf{Visualizations of parallel executions and cooperative behaviors.} In the first example, there are three feasible subtasks. The \textcolor{blue}{blue} agent chooses the subtask $\emph{Wash(Dish)}$ because the dirty plate is both reachable from and close to him. The \textcolor{magenta}{magenta} agent chooses the subtask $\emph{Pick(Onion)}$ because the onion supply is reachable from him. In the second example, the two agents cooperate on the subtask $\emph{Chop(Onion)}$. The \textcolor{magenta}{magenta} agent picks the onion and passes it to the \textcolor{blue}{blue} agent, while the  \textcolor{blue}{blue} agent finalizes the subtask by chopping the onion at the chopping block.}
	\label{fig:more vis}
\end{figure}

Figure~\ref{fig:more vis} shows more detailed visualizations of two emergent behaviors.

\subsection{Baseline Description}
\label{app:baseline}
In our study, we mainly consider two types of baselines: state-of-the-art MARL algorithms and natural language-guided agents.
\begin{itemize}
    \myitem State-of-the-art multi-agent reinforcement learning. We directly use MAPPO~\cite{yu2021surprising} to train joint policies without the guiding program, while still providing dense rewards to the agents when any correct subtask is completed; 
    \myitem Natural language-guided agent. We train a goal-conditioned policy where the goal is encoded from the natural language description with a pre-trained BERT model in the PyTorch package \textit{transformer}~\citep{devlin2018bert}. The encoded features of the tokens are average-pooled and frozen during the training process. Then we use a learnable MLP to encode the frozen features into goal features. The MLP has three layers connected by the ReLU activation, and the hidden sizes of the MLP are all 128. We note that here we do not provide a program-like structure for the language-guided agents.
\end{itemize}
The baseline models have the same policy and value network architecture as those of E-MAPP. When training the end-to-end baseline models, the agents are rewarded $0.2$ if they complete a correct subtask and $1$ if the final task is completed. We also use self-imitation learning in baseline algorithms to address the sparse reward problem.

The computation costs of \model and baselines are shown in Table~\ref{tab:computation  cost}

\begin{table}[h]
  \caption{\textbf{Computation Cost.} The number of parameters and the running time in \model and baselines. }
  \label{tab:computation cost}
  \centering
\begin{tabular}{ccc}
\toprule[0.5mm]
     Model & Parameters & Running Time \\ \midrule 
     \model & 5.6M & around 72h\\
     natural language guided agent & 2.3M & around 36h \\
     MAPPO & 1.8M & around 24h
\\ \bottomrule
\end{tabular}
\end{table}

\subsection{Results in the Stacking Environment}
\label{app:robot}
\begin{figure}[htbp]
	\centering
    \fbox{\includegraphics[trim=0cm 4cm 0cm 0cm, width=\textwidth]{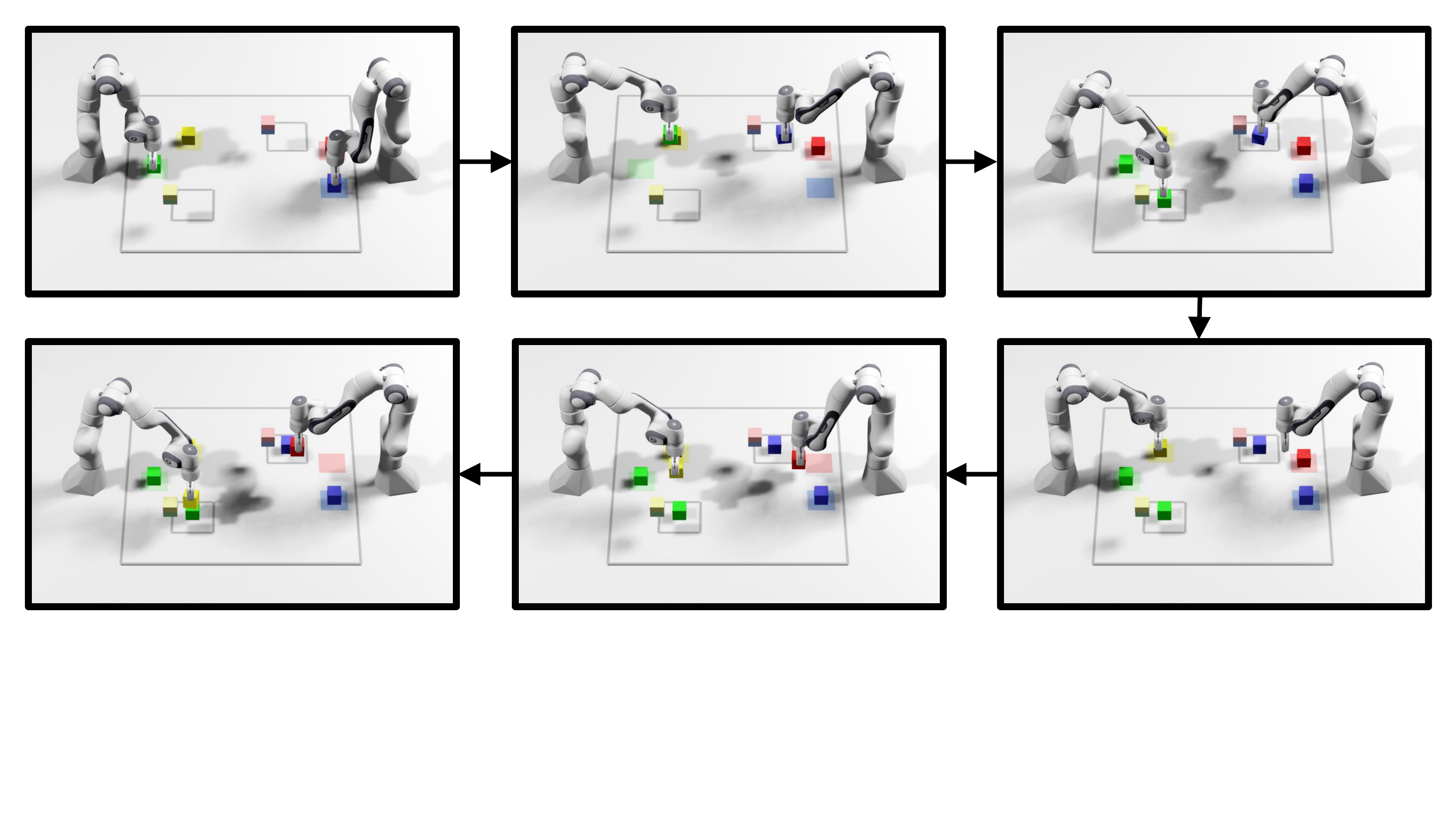}}
	\caption{\textbf{A demonstration of the completion process of stacking tasks.} The pictures show the process of picking up boxes and stacking them.}
	\label{fig:robot}
\end{figure}

To evaluate the ability of \model to scale to more complex control tasks, we also demonstrate how \model works on a parallel stacking task set. It is desired that the two franka arms cooperate to complete two stacks of blocks in a given order. The behavior primitives considered in this setting are in the form of \textit{Stack($c$, $idx$)}, which represents the subtask of putting a $c$-colored block on the top of the $idx$-th pile. We use motion planning as the policy to complete the subtasks. Figure~\ref{fig:robot} is a demonstration of the task completion process. The baselines considered above fail to achieve the goal in any episode, while \model can achieve a 46\% completion rate.

\subsection{Computational Resources}
We train our model on a single Nvidia TITAN-X GPU, in a 16-core Ubuntu 18.04 Linux server.

\subsection{Examples of Programs Used in Evaluation}
\subsubsection{Easy tasks}
\begin{verbatim}
if IsOnFire(): 
    PutOutFire()
\end{verbatim}
\begin{verbatim}
if is_ordered(ChoppedTomato): 
    Serve(ChoppledTomato+Plate)
\end{verbatim}
\begin{verbatim}
if is_ordered(ChoppedOnion): 
    Pick(FreshOnion)
\end{verbatim}
\subsubsection{Medium tasks}
\begin{verbatim}
repeat:
    Pick(FreshTomato)
\end{verbatim}
\begin{verbatim}
parallel:
    1. Pick(FreshOnion)
    2. Pick(FreshTomato)
    3. WashDirtyPlate()
\end{verbatim}
\begin{verbatim}
if is_ordered(ChoppedTomato):
    Chop(FreshTomato)
    Merge(ChoppedTomato,Plate)
    Serve(ChoppedTomato+Plate)
\end{verbatim}
\subsubsection{Hard tasks}
\begin{verbatim}
parallel:
    1: if is_ordered(Onion):
            Merge(ChoppedOnion,Plate)
            Serve(ChoppedOnion) 
    2: if is_ordered(Onion):
            Merge(ChoppedTomato,Plate)
            Serve(ChoppedTomato)
    3: while (True):
        If (IsOnFire()):
             PutOffFire() 
\end{verbatim}
\begin{verbatim}
parallel:
        1: 
            Pick(FreshOnion)
            Chop(FreshOnion)
        2:
            Pick(FreshTomato)
            Chop(FreshTomato)
        3:
            WashDirtyPlate()
        4:
            Merge(ChoppedOnion,Plate)
            Serve(ChoppedOnion)
        5:
            Merge(ChoppedTomato,Plate)
            Serve(ChoppedTomato)
\end{verbatim}
\subsection{More experiments}

\subsubsection{Overcooked with more agents}
We conduct an experiment with a doubled number of agents to evaluate our algorithm. Table ~\ref{tab:more agents} shows the results. The results indicate that E-MAPP can scale to environments with more agents and further boost the time efficiency by parallelization.

\begin{table}[h]
  \caption{\textbf{Additional experiment in Overcooked involving four agents.} E-MAPP can scale to environments with more agents and further boost the time efficiency by parallelization. }
  \label{tab:more agents}
  \centering
\begin{tabular}{ccc}
   \toprule[0.5mm]
   model & score & completion rate  \\  
   \midrule
   E-MAPP(original) & 0.99±0.22 & 43.7\%\\
   E-Mapp(larger) & 1.13± 0.31 & 46.3\% \\
\bottomrule
\end{tabular}
\end{table}
\subsubsection{Comparison with other centralized execution agents}
We also compare E-MAPP with a centralized execution approach. We implement a centralized PPO where joint policy is directly produced by a centralized network. Table ~\ref{tab:centralized ppo} shows the results. The results indicate that the centralized PPO suffers from the high dimensionality of the joint action space and fails to learn cooperation and coordination. 
\begin{table}[h]
  \caption{\textbf{Comparison of E-MAPP, centralized algorithm PPO and decentralized algorithm MAPPO.}  We additionally compare E-MAPP with a centralized PPO that directly outputs the joint policy.}
  \label{tab:centralized ppo}
  \centering
\begin{tabular}{cc}
   \toprule[0.5mm]
   model & score\\  
   \midrule
   E-MAPP &  1.58±0.60 \\
   MAPPO(decentralized) &  0.59± 0.27\\
   MAPPO(centralized) &  0.48 ± 0.21\\
\bottomrule
\end{tabular}
\end{table}

\subsection{Potential parallel-program synthesis approaches}
\label{app:program synthesis}
The guiding program in our work can be obtained with program synthesis approaches. When it comes to a new domain, we can devise new perception primitives and behavior primitives based on object properties and interactions among objects~\citep{liu2019learning}. These primitives, along with the branching and parallelization keywords, compose the DSL. Previous approaches on program synthesis~\citep{sun2018neural,devlin2017robustfill,chen2018execution,chen2021latent} can be applied to synthesize programs for tasks. For example, we can synthesize programs from diverse video demonstrations. The activities~(subtasks) of a task in a video can be segmented out as a subroutine for program extraction~\citep{sun2018neural}. By summarizing the chronological order of subtask completions, we can obtain the dependence of subtasks and put the possibly parallelizable subtask in one parallel subroutine.

\end{document}